# Increasing Compression Ratio in PNG Images by k-Modulus Method for Image Transformation

**Firas A. Jassim**
Faculty of Administrative Sciences
Management Information Systems Department
Irbid National University
Irbid, 2600, Jordan
**firasajil**@yahoo.com

**Abstract**
Image compression is an important filed in image processing. The science welcomes any tinny contribution that may increase the compression ratio by whichever insignificant percentage. Therefore, the essential contribution in this paper is to increase the compression ratio for the well known Portable Network Graphics (PNG) image file format. The contribution starts with converting the original PNG image into k-Modulus Method (k-MM). Practically, taking k equals to ten, and then the pixels in the constructed image will be integers divisible by ten. Since PNG uses Lempel-Ziv compression algorithm, then the ability to reduce file size will increase according to the repetition in pixels in each k×k window according to the transformation done by k-MM. Experimental results show that the proposed technique (k-PNG) produces high compression ratio with smaller file size in comparison to the original PNG file.

**Keywords:** Image compression, PNG, compression ratio, PSNR.

## 1. INTRODUCTION

Portable Network Graphics (PNG) is an image file format for storing, transmitting, and displaying images. In fact, PNG was firstly designed to replace both Graphics Interchange Format (GIF) and Tagged Image File Format (TIFF) image formats in many applications [9] . PNG is treated as a lossless image compression with transparency information [2] . PNG also supports more advanced attributes such as gamma correction [6] . However, PNG was never intended to compete with JPEG on its own terms. But PNG, like GIF, is more appropriate than JPEG for images with few colors or with lots of sharp edges, such as cartoons or bitmapped text. PNG also provides direct support for gamma correction [5] [9] . Since PNG compression is completely lossless and since it supports up to 48-bit truecolor or 16-bit grayscale. Hence, saving, restoring, and resaving an image will not impair its quality. On the contrary, standard Joint Graphics Expert Group (JPEG) will degrade image quality during the compression process [13] . Moreover, PNG also supports full transparency information; unlike JPEG that has no transparency at all While GIF has no partial transparency. Concerning TIFF, the full transparency option is part of the process being used but is not an essential demand [6] [14] .

Dictionary based schemes (such as ZIP) are widely used for computed graphic image compression on the Internet (such as GIF, TIFF, and PNG) [1] . PNG uses the Deflate compression method that is used in the widespread ZIP file format [2] . In fact, deflate is an enhanced version of the Lempel-Ziv compression algorithm [7] . It acts similarly to the LZW algorithm, and searches for reiterated horizontal patterns along each scan line. Moreover, to enhance the compression process, PNG pre-scan the image data using some prediction functions before the compression process. Hence, PNG and GIF share the same technique by compressing horizontal patterns, but PNG has an additional step which is also scans the vertical patterns [8] . However, the horizontal and vertical patterns scan gives the powerful for PNG especially with solid colors that are the basic elements in cartoon images. Furthermore, it must be avoided bring in detail or noise into PNG images. Also, in PNG compression, it must be avoided to shudder and dithering; it fractures tabular areas of color, and makes PNG compression less efficient [14] .
According to JPEG process, some images are not ideal for JPEG compression. Actually, JPEG is excellent for images where there are high details. However, the compression techniques are not as well suited for images where there are straight lines that divide flat areas of single color. Basically, such images contain diagrams, art with text can not displayed exactly by JPEG. Therefore, the best compression technique that best fit these images is the PNG compression [5] . JPEG compression of images is greatly effective at reducing file size of some images with minimal loss of printed image quality. In the contrary, PNG compression of images with large blocks of single color maybe both more efficient and greater in printed quality [6] [11] .

## 2. k-MODULUS METHOD

The first appearance of k-Modulus method (k-MM) was the Five Modulus Method (FMM) that firstly originated by [3] . Basically, it was initiated as a transformation method for image compression. The essential concept of FMM was to transform the entire image pixels into multiple of five. Lately, the k-Modulus Method (k-MM) was founded as a generalization for the FMM. In fact, the basic idea behind k-MM is to transform the whole image into multiples of k, where k is any integer between 2 to 25 [4] . Since the human eye does not differentiate between the original image and the transformed k-MM image [4] . It is known that, for each of the R, G, and B arrays in the color image are consist of pixel values varying from 0 to 255. The main impression of k-MM is to transform each pixel inside the original image into integers divisible by k. Here, in this paper, the best value of k that keeps the transformed image undetectable is 10. Therefore, by transforming the whole pixels inside the image into multiples of 10, this will gives two benefits. The first benefit is that making all pixels inside k×k block have similar values with low dispersion between pixels. On the other hand, the second benefit is that, this process is suitable for images with few colors as in cartoon image which is the aim of this paper. According to figure (1), it is clear that the k-MM (k=10) transformation does not affect the Human Visual System (HVS).

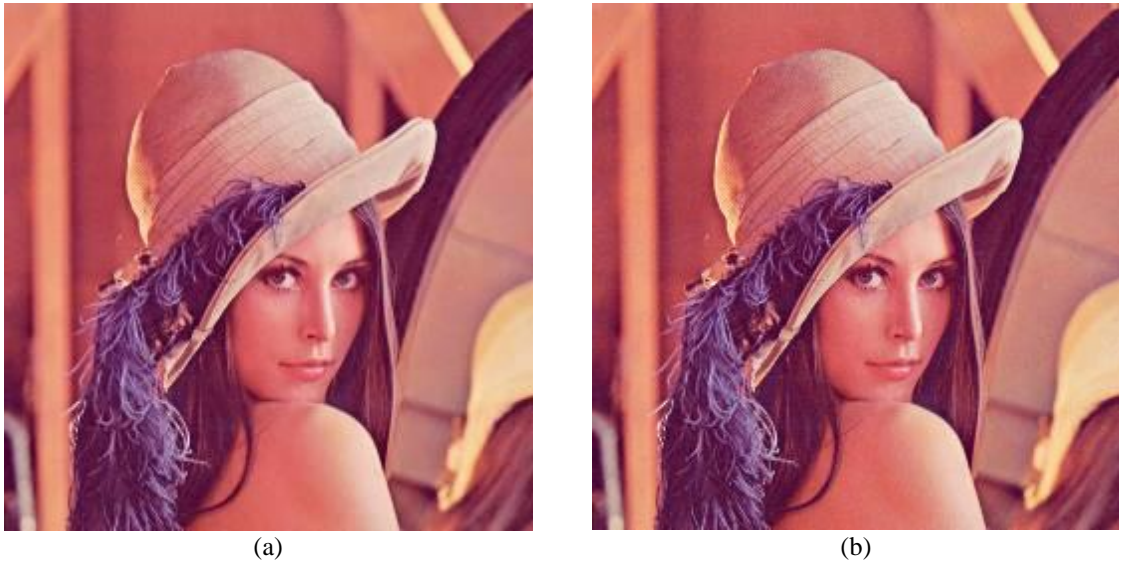

(a) (b)
**FIGURE 1:** (A) Original BMP image (B) Transformed image with 10-Modulus Method

According to figure (1), the visual differences between the original image and the transformed (k-MM) image are undistinguishable. It must be mentioned that when increasing k more than 10, this will contaminate some parts of the image. Therefore, the most affordable k that keeps the differences as small as possible is 10. Mathematically speaking, an illustrative example will be introduced to explain the k-MM process exactly. In accordance with figure (2), a 10×10 block has been taken from the original bitmap Lena image.

| 141 | 128 | 107 | 84  | 81  | 81  | 112 | 136 | 133 | 72  |
|-----|-----|-----|-----|-----|-----|-----|-----|-----|-----|
| 122 | 106 | 86  | 80  | 92  | 107 | 134 | 140 | 113 | 67  |
| 117 | 98  | 80  | 76  | 108 | 138 | 145 | 137 | 79  | 66  |
| 98  | 92  | 80  | 83  | 121 | 154 | 146 | 130 | 68  | 64  |
| 87  | 81  | 76  | 92  | 137 | 156 | 148 | 90  | 70  | 83  |
| 81  | 74  | 72  | 99  | 139 | 147 | 132 | 72  | 100 | 132 |
| 87  | 73  | 76  | 107 | 138 | 144 | 126 | 103 | 148 | 162 |
| 87  | 77 | 86  | 133 | 144 | 139 | 135 | 152 | 178 | 179 |
| 92  | 79 | 108 | 142 | 144 | 134 | 150 | 184 | 201 | 185 |
| 114 | 89 | 124 | 145 | 145 | 128 | 152 | 197 | 184 | 172 |

**FIGURE 2:** 10×10 Block from Lena.bmp

After applying the k-MM with k equals to 10. The resulted 10×10 block have been presented in figure 3.

| 140 | 130 | 110 | 80  | 80  | 80  | 110 | 140 | 130 | 70  |
|-----|-----|-----|-----|-----|-----|-----|-----|-----|-----|
| 120 | 110 | 90  | 80  | 90  | 110 | 130 | 140 | 110 | 70  |
| 120 | 100 | 80  | 80  | 110 | 140 | 140 | 140 | 80  | 70  |
| 100 | 90  | 80  | 80  | 120 | 150 | 150 | 130 | 70  | 60  |
| 90  | 80  | 80  | 90  | 140 | 160 | 150 | 90  | 70  | 80  |
| 80  | 70  | 70  | 100 | 140 | 150 | 130 | 70  | 100 | 130 |
| 90  | 70  | 80  | 110 | 140 | 140 | 130 | 100 | 150 | 160 |
| 90  | 80  | 90  | 130 | 140 | 140 | 130 | 150 | 180 | 180 |
| 90  | 80  | 110 | 140 | 140 | 130 | 150 | 180 | 200 | 180 |
| 110 | 90  | 120 | 140 | 140 | 130 | 150 | 200 | 180 | 170 |

**FIGURE 3:** 10×10 Block from transformed Lena.bmp using k-MM

Clearly, the differences between the original 10×10 block and the transformed block are negligible, figure 4.

| 1  | -2 | -3 | 4  | 1  | 1  | 2  | -4 | 3  | 2  |
|----|----|----|----|----|----|----|----|----|----|
| 2  | -4 | -4 | 0  | 2  | -3 | 4  | 0  | 3  | -3 |
| -3 | -2 | 0  | -4 | -2 | -2 | 5  | -3 | -1 | -4 |
| -2 | 2  | 0  | 3  | 1  | 4  | -4 | 0  | -2 | 4  |
| -3 | 1  | -4 | 2  | -3 | -4 | -2 | 0  | 0  | 3  |
| 1  | 4  | 2  | -1 | -1 | -3 | 2  | 2  | 0  | 2  |
| -3 | 3  | -4 | -3 | -2 | 4  | -4 | 3  | -2 | 2  |
| -3 | -3 | -4 | 3  | 4  | -1 | 5  | 2  | -2 | -1 |
| 2  | -1 | -2 | 2  | 4  | 4  | 0  | 4  | 1  | 5  |
| 4  | -1 | 4  | 5  | 5  | -2 | 2  | -3 | 4  | 2  |

**FIGURE 4:** differences between the original and transformed 10×10 blocks

## 3. PROPOSED TECHNIQUE

The proposed contribution in this paper comes from the ability to construct a novel format to reduce the PNG file size. Basically, the novel format is a lossy format while the original PNG format is a lossless format. In another words, the ability to restore the original file in the original PNG format is 100%, while in the proposed PNG is high but not 100%. As mentioned in the section 2, the proposed method is suitable for cartoon or low details images only. The proposed technique starts with converting the original R, G, and B arrays in the original PNG file via k-Modulus Method (KMM). According to [4] , the best k that keeps the visual differences quite acceptable is 10. Hence, by using 10-MM, the constructed image has very high similarity to the original PNG file. The novel image format has greater compression ratio and this has been obtained with two factors. The firs one is that, the original PNG file uses LZW77 approach in it compression process [2] . The second one is that the k-Modulus Method has small dispersion between the k×k block pixels. Therefore, merging these two factor produce an approximately exact copy of the original image that can not be distinguished easily by the human eye. The proposed technique has some pros and cons, one of the best pros is the high compression ratio but on the other hand there are some cons. One of these cons is the distortion in some of the gradient parts, i.e. the proposed technique does not fit images that have any gradient parts. Hence, as mentioned previously, the proposed technique is best suitable for cartoon images.

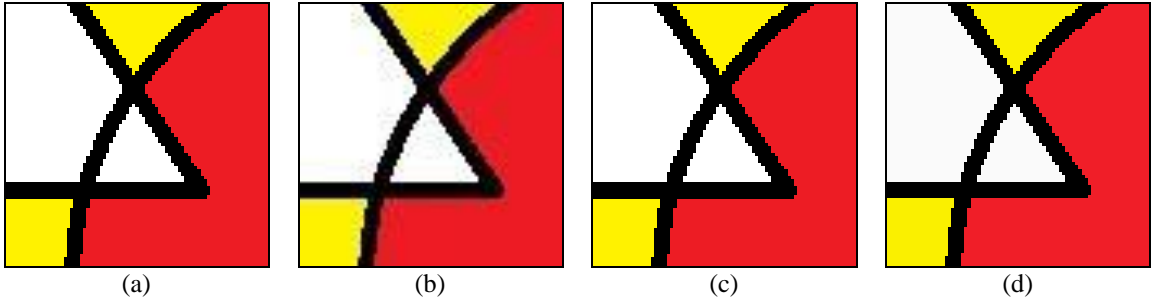
**FIGURE 5:** (A) BMP (B) JPEG (C) PNG (d) FPNG

## 4. EXPERIMENTAL RESULTS

In order to support the proposed contribution in this paper, a variety of test image have been used, figure (6). The constructed k-PNG images have absolutely undistinguishable ocular differences by the human eye. But this is not enough, therefore; a mathematical performance measures have been applied. The first and the simplest evaluation measure is the Mean Square Error (MES) [13] . The second one is the widely used measure in compression process which is the Peak Signal-to-Noise Ratio (PSNR) [12] . The last measure is the structure similarity index (SSIM) which was proposed by [10] [16] . All the three performance measures (MSE, PSNR, and SSIM) have been computed between the original PNG and the proposed technique (k-PNG). First of all, there must be a comparison by file sizes between each of BMP, PNG, and k-PNG, table (1). The k-PNG file size is the smallest. This is because the PNG format can take advantage of the low number of colors and the large areas of single color to greatly reduce file size. According to table (2), and in accordance with MSE, the MSE values are too small which means that there are completely small differences between the original PNG and k-PNG. Moreover, concerning PSNR (table 2), the PSNR values are too high which means that the proposed compression process is quite acceptable in accordance with [15] . Finally, the very high values of SSIM (table 2), show that there are high similarity between the original PNG and the proposed k-PNG.

Furthermore, from the compression ratio consideration, a comparison between the compression ratios between the original PNG and the proposed k-PNG, each was compared with the original Bitmap image (BMP). Numerically the results were introduces in table (1) and graphically in figure (7). Concerning SHAPES image, the compression ratio for the original PNG was (1:69) while in the proposed was found to be (1:183) which is very high compression ratio. Just for the matter of similarity and comparison, JPEG compression must be mentioned. Therefore, SHAPES images was compressed using JPEG and the file size was (26.2 KB) with compression ratio (29.3). Clearly, this indicates that k-PNG format produces high compression ratio which is greater than JPEG but in special images only. This contradicts the widely known concept that claims that JPEG is the optimal compression ratio ever reached. Once again, it must be mentioned that, the proposed k-PNG is suitable for cartoon images or images with few colors in flat regions.

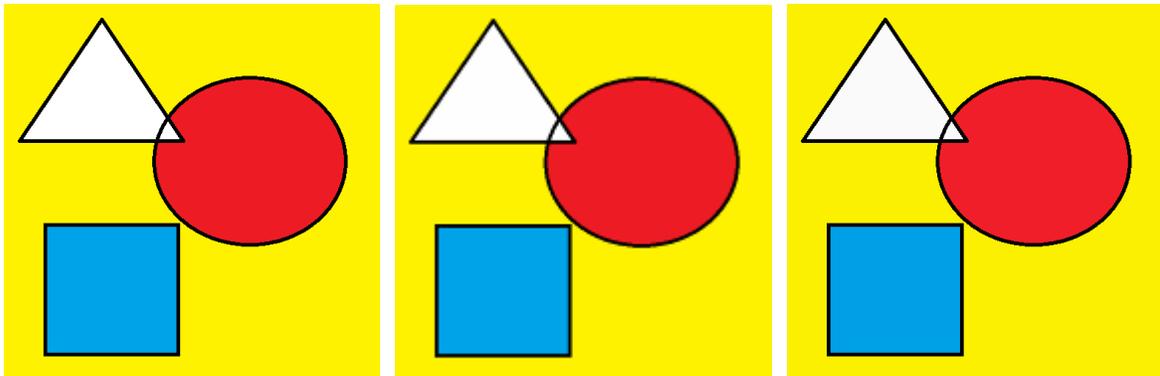

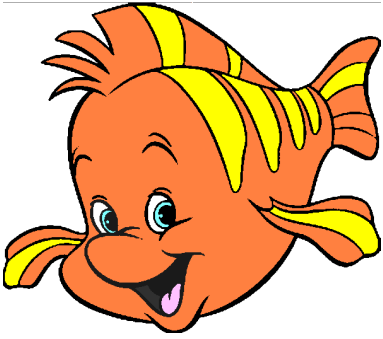 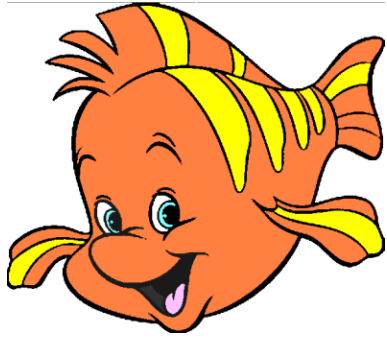 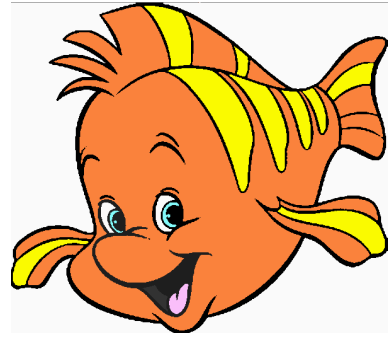
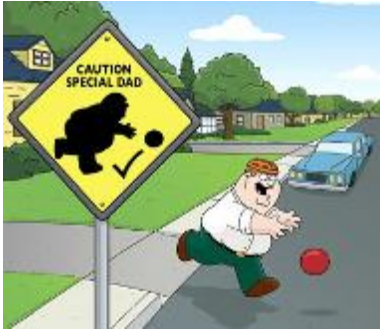 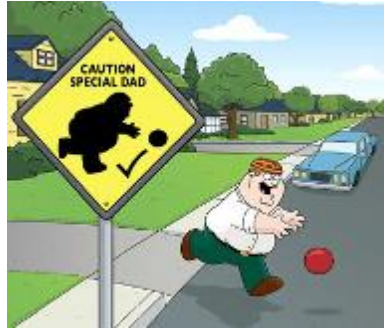 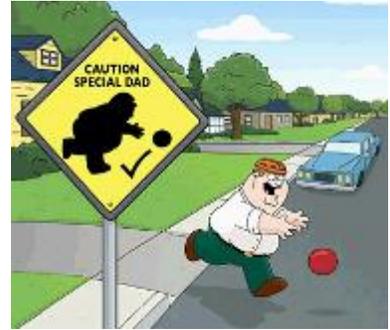
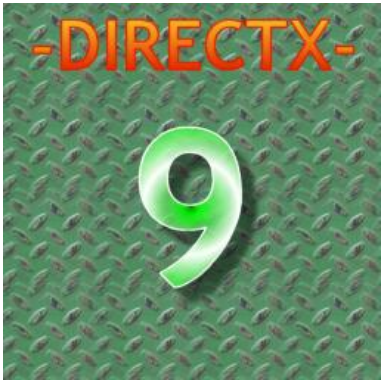 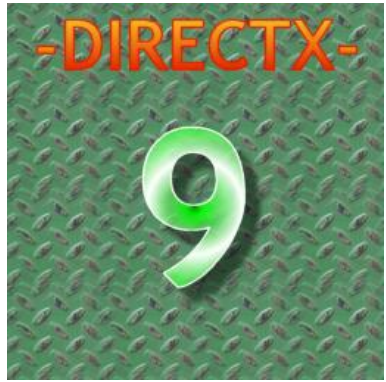 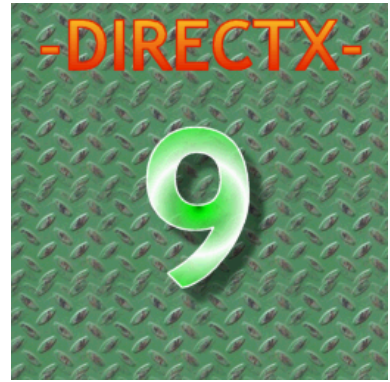
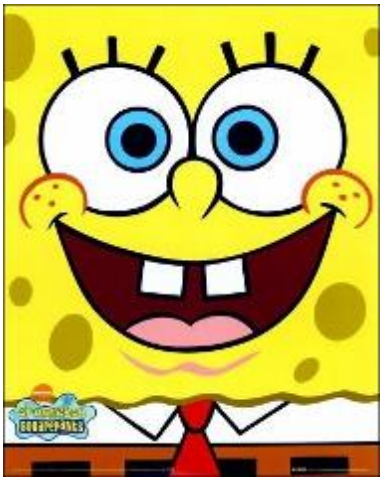 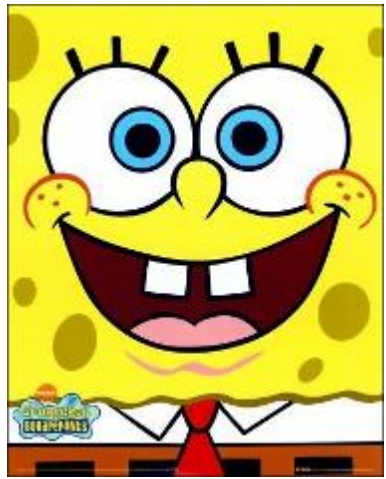 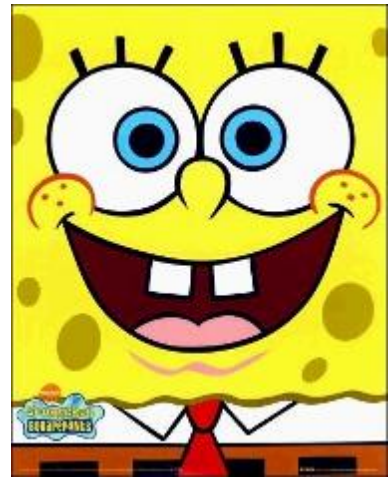

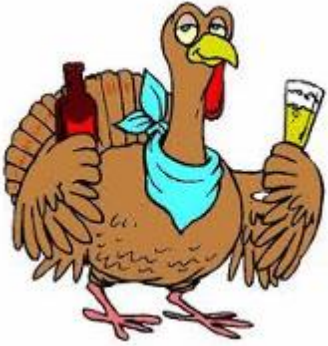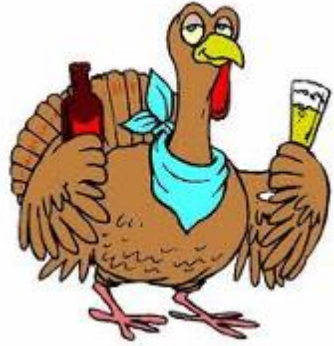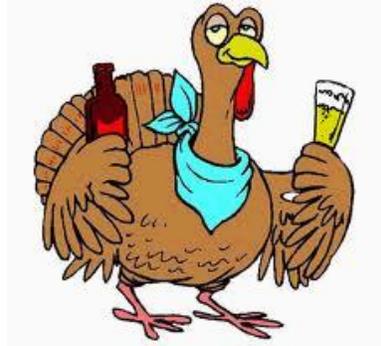
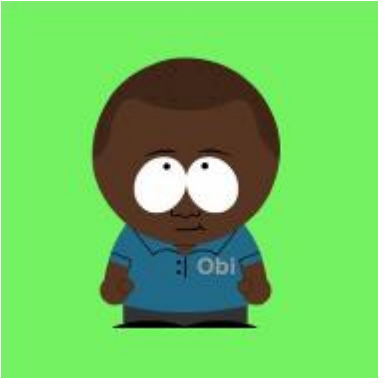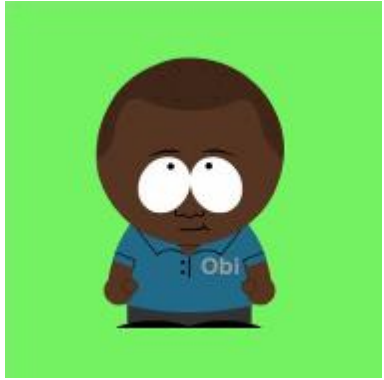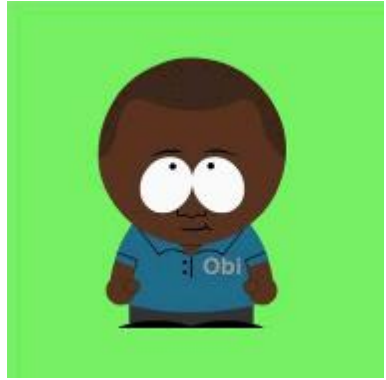
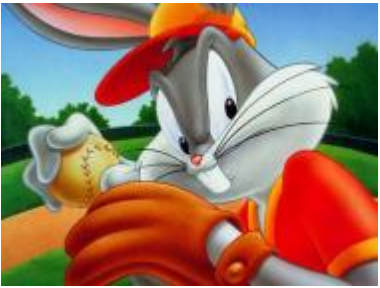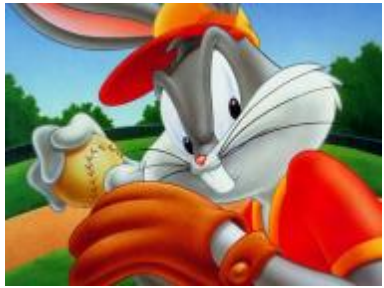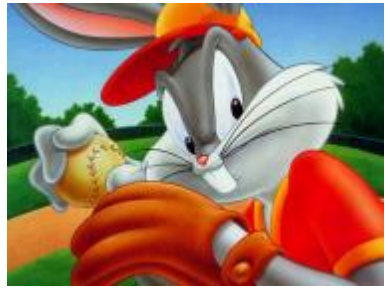
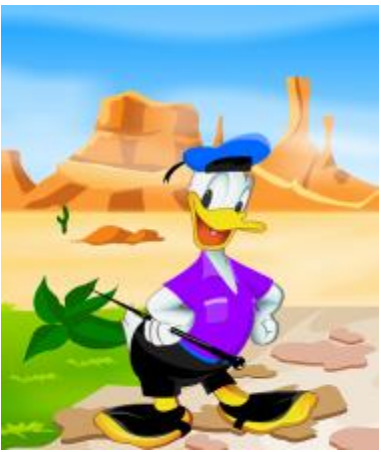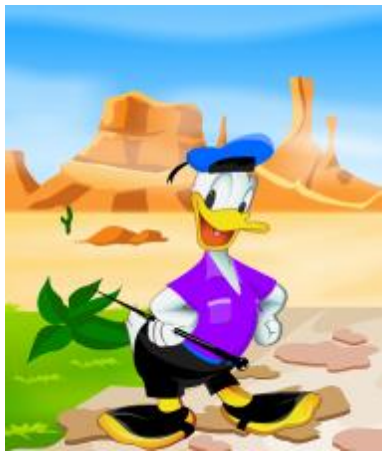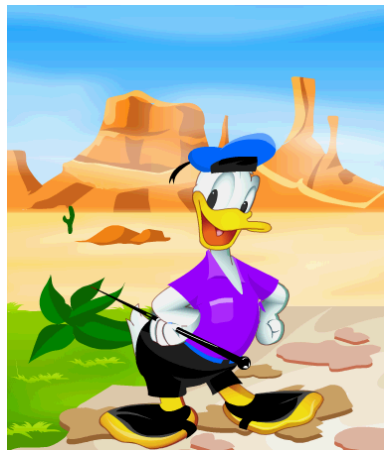

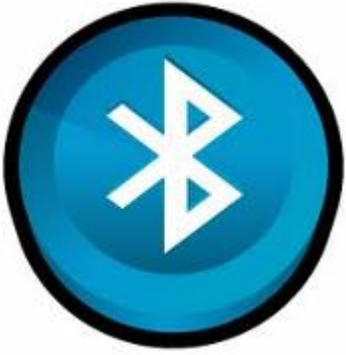 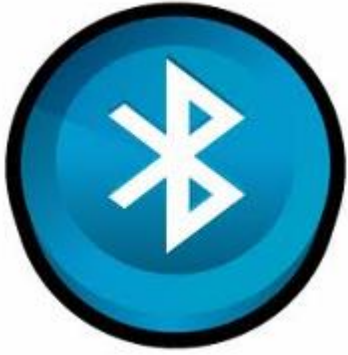 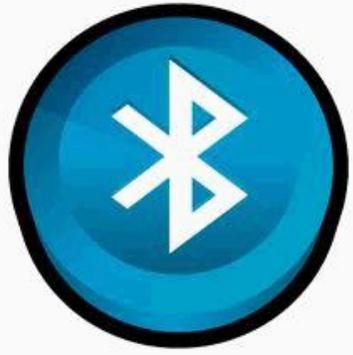
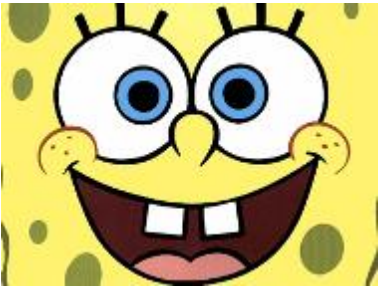 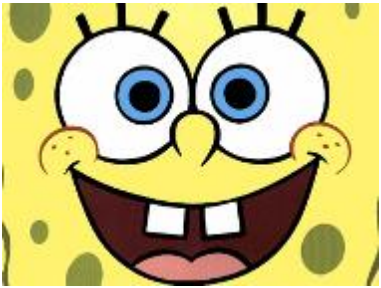 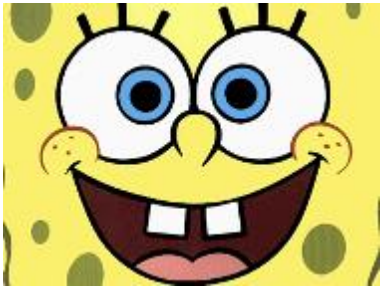
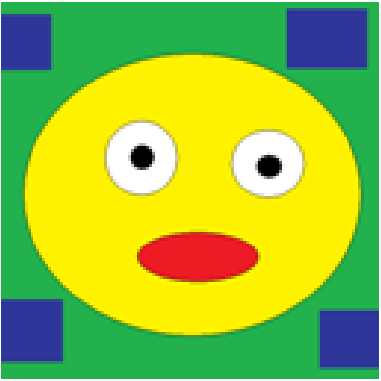 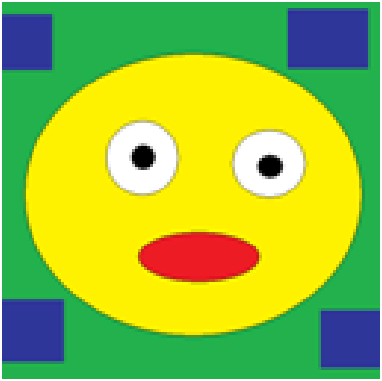 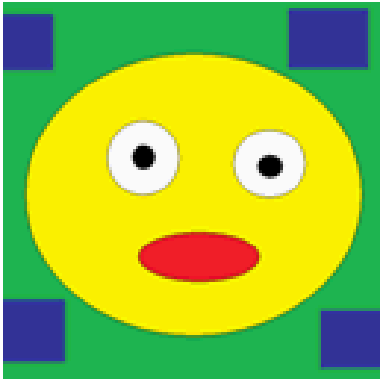
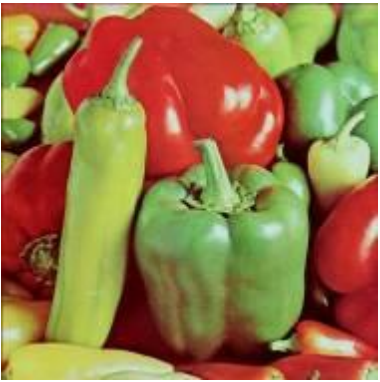 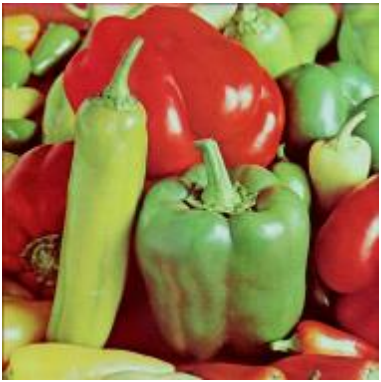 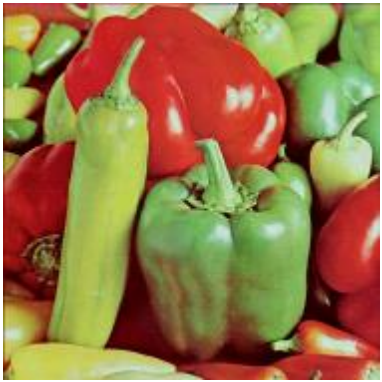

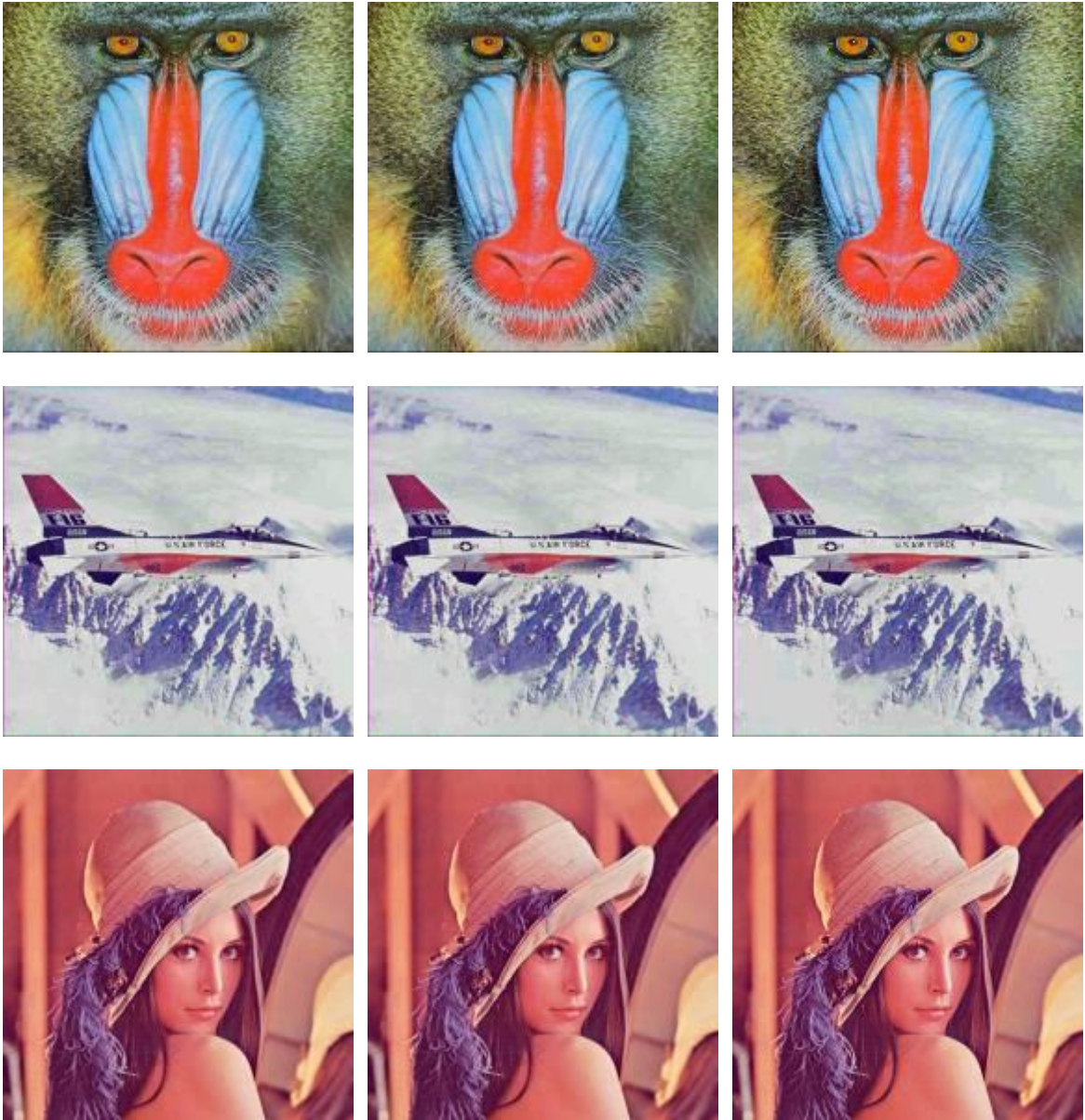

**FIGURE 6:** Variety of test images (first column) BMP (second column) PNG (third column) FPNG

|  | Dimensions | BMP | PNG | | k-PNG | |
|---|---|---|---|---|---|---|
|  |  | File size | File size | CR | File size | CR |
| **Shapes** | 512×512 | 768 KB | 11.1 KB | 69.2 | 4.19 KB | 183.3 |
| **Fish** | 599×525 | 922 KB | 25.2 Kb | 36.6 | 14.9 KB | 61.9 |
| **Caution** | 360×313 | 330 KB | 186 KB | 1.8 | 79.6 KB | 4.1 |
| **Direct x** | 256×256 | 192 KB | 81.8 KB | 2.3 | 33.2 KB | 5.8 |
| **Sponge1** | 360×452 | 476 KB | 284 KB | 1.7 | 88.2 KB | 5.4 |
| **Turkey** | 225×225 | 148 KB | 83.0 KB | 1.8 | 37.2 KB | 4.0 |
| **Obi** | 385×385 | 434 KB | 85.9 KB | 5.1 | 29.3 KB | 14.8 |
| **Bunny** | 1152×864 | 2.84 MB | 1.99 MB | 1.4 | 429 KB | 6.6 |
| **Duck** | 379×449 | 499 KB | 107 KB | 4.7 | 44.9 KB | 11.1 |
| **Bluetooth** | 225×225 | 148 KB | 69.7 KB | 2.1 | 22.1KB | 6.7 |
| **Sponge2** | 350×262 | 269 KB | 115 KB | 2.3 | 39.6 KB | 6.8 |
| **Face** | 128×128 | 48.0 KB | 7.44 KB | 6.5 | 4.16 KB | 11.5 |
| **Lena** | 512×512 | 768 KB | 465 KB | 1.7 | 232 KB | 3.3 |
| **Peppers** | 512×512 | 768 KB | 376 KB | 2.0 | 166 KB | 4.6 |
| **Baboon** | 512×512 | 768 KB | 613 KB | 1.3 | 381 KB | 2.0 |
| **F16** | 512×512 | 768 KB | 225 KB | 3.4 | 113 KB | 6.8 |

**TABLE 1:** Comparison of file sizes and compression ratio between BMP, PNG, and k-PNG

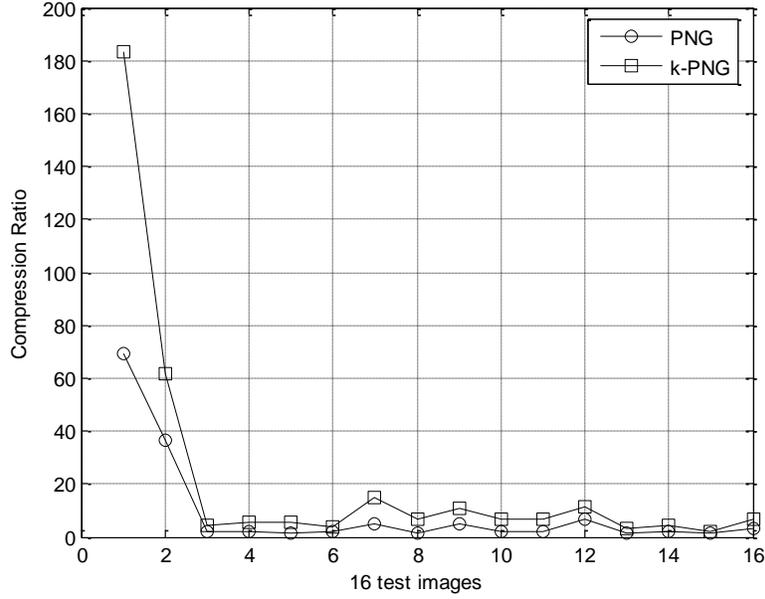

**FIGURE 7:** (A) BMP (B) JPEG (C) PNG (d) k-PNG

| | MSE | PSNR | SSIM |
|---|---|---|---|
| **Shapes** | 5.3578 | 40.8410 | 0.9998 |
| **Fish** | 8.6882 | 38.7415 | 0.9997 |
| **Caution** | 3.5114 | 42.6760 | 0.9855 |
| **Direct x** | 2.8248 | 43.6209 | 0.9877 |
| **Sponge1** | 4.1892 | 41.9095 | 0.9852 |
| **Turkey** | 10.1427 | 38.0693 | 0.9952 |
| **Obi** | 1.5427 | 46.2479 | 0.9914 |
| **Bunny** | 2.9083 | 43.4945 | 0.9676 |
| **Duck** | 2.9592 | 43.4191 | 0.9859 |
| **Bluetooth** | 8.1995 | 38.9929 | 0.9826 |
| **Sponge2** | 6.7009 | 39.8695 | 0.9920 |
| **Face** | 3.8891 | 42.2323 | 0.9983 |
| **Lena** | 2.9264 | 43.4674 | 0.9714 |
| **Peppers** | 2.8515 | 43.5800 | 0.9717 |
| **Baboon** | 2.9063 | 43.4974 | 0.9905 |
| **F16** | 4.5831 | 41.5192 | 0.9772 |

**TABLE 2:** MSE, PSNR, and SSIM for 16 test images

## 5. CONCLUSION

The basic conclusion in this paper is that, k-PNG is a novel image format with high compression ratio compared with the original PNG file format and also with small file size. The proposed image file format is suitable for cartoon images or images with low number of colors in large areas of single color. It must be mentioned that, the proposed technique (k-PNG) is not considered in this paper as the optimal lossless image compression technique but indeed it may antiquates the original PNG image when there is a need for moderate quality in images with small files size at the same time. Consequently, the proposed technique is very suitable for the transmission especially over the World Wide Web since it compress the file size less than PNG. Finally, the proposed technique does not fit images that have any gradient parts because this will produce distortion in these parts.

## 6. REFERENCES

[1] D. A. Clunie, "Lossless Compression of Grayscale Medical Images - Effectiveness of Traditional and State of the Art Approaches", In Proceedings of SPIE (Medical Imaging), Vol. 3980, February 2000.

[2] D. Salomon, *Data Compression: The Complete Reference*, Fourth Edition, Springer.

[3] F. A. Jassim, "Five Modulus Method for Image Compression", Signals and Image Processing: An International Journal (SIPIJ), Vol.3, No.5, pp. 19-28, October 2012.

[4] F. A. Jassim, "k-Modulus Method for Image Transformation", International Journal of Advanced Computer Science and Applications (IJACSA), Vol. 4, No. 3, pp. 267-271, 2013.

[5] G. Roelofs, PNG: The Definitive Guide, O'Reilly Media, 1999.

[6] http://www.libpng.org

[7] J. Ziv. and A. Lempel, "A Universal Algorithm for Sequential Data Compression," IEEE Transactions on Information Theory, Vol. 23, No. 3, pp. 337-343, 1977.

[8] K. Sayood, *Introduction to Data Compression*, Third edition, Morgan Kaufmann Publishers.

[9] L. Crocker, "PNG: The Portable Network Graphic Format", Dr. Dobb's Journal, July 1995. pp. 36-44.

[10] L. Zhang, L. Zhang, X. Mou and D. Zhang, "FSIM: A Feature SIMilarity Index for Image Quality Assessment", IEEE Transactions on Image Processing, Vol. 20, No. 8, pp. 2378-2386, 2011.

[11] M. Domanski, Krzysztof Rakowski, "Lossless And Near-Lossless Image Compression With Color Transformations", Proceedings 2001 International Conference on Image Processing ICIP, Thessaloniki October 7-10 2001. Vol. III pp. 454-457.

[12] Q. Huynh-Thu, M. Ghanbari, "Scope of validity of PSNR in image/video quality assessment". Electronics Letters, Vol. 44, No. 13, pp.800–801, 2008.

[13] R. C. Gonzalez R. C. and R. E. Woods, "Digital Image Processing", Second edition, 2004.

[14] T. Boutell and G. Randers-Pehrson, PNG (Portable Network Graphics) Specification, The latest PNG specification. W3C Tech Reports, 2003.

[15] T. Welstead Stephen. Fractal and wavelet image compression techniques, SPIE Publication, pp.155–156, 1999.

[16] Z. Wang., A. C. Bovik, "Image Quality Assessment: From Error Visibility to Structural Similarity", IEEE Transactions on Image Processing, Vol. 13, No. 4, pp. 600-612, 2004.